# CAMERA-BASED METHOD FOR THE DETECTION OF LIFTED TRUCK AXLES USING CONVOLUTIONAL NEURAL NETWORKS

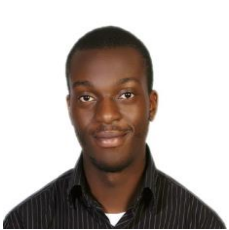 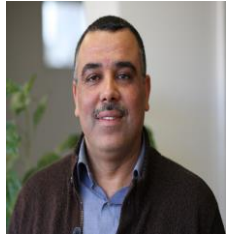 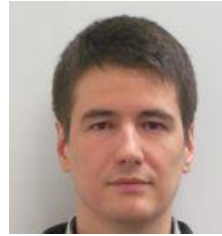 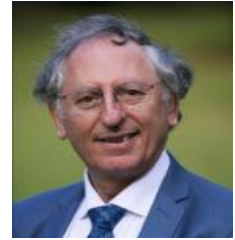

**B. TCHANA TANKEU**
Research Engineer in signal processing at Cerema, Bron, France. Expertise in high resolution signal processing methods and non-destructive pavement survey. Obtained his Ph.D. in signal and image processing from the University of Nantes, France in 2023.

**M. BOUTELDJA**
Researcher and Project Manager at Cerema Bron, France. Expertise in heavy vehicle dynamic, data processing & analysis, road pavement and safety. Obtained his Ph.D. in Automatics and Robotics from Versailles Saint Quentin-en-Yvelines University (France), in 2005.

**N. GRIGNARD**
Head manager in Infrastructure & Skid resistance at Cerema, Bron, France. Expertise in truck weigh-in-motion algorithms, data processing & analysis, road monitoring. Graduated from the national school of public works (Ecole nationale des travaux publics de l'état) in 2007.

**B. JACOB**
Honorary general engineer of bridges and roads, emeritus research adviser at Gustave Eiffel University, France. Expert in Road Freight Transport, Bridge Loading and Weigh-in-Motion. Graduated from Ecole Polytechnique and Ecole Nationale des Ponts et Chaussées.

**Abstract**

The identification and classification of vehicles play a crucial role in various aspects of the control-sanction system. Current technologies such as weigh-in-motion (WIM) systems can classify most vehicle categories but they struggle to accurately classify vehicles with lifted axles. Moreover, very few commercial and technical methods exist for detecting lifted axles. In this paper, as part of the European project SETO (Smart Enforcement of Transport Operations), a method based on a convolutional neural network (CNN), namely YOLOv8s, was proposed for the detection of lifted truck axles in images of trucks captured by cameras placed perpendicular to the direction of traffic. The performance of the proposed method was assessed and it was found that it had a precision of 87%, a recall of 91.7%, and an inference time of 1.4 ms, which makes it well-suited for real time implantations. These results suggest that further improvements could be made, potentially by increasing the size of the datasets and/or by using various image augmentation methods.

## 1. Introduction

For several years now, the demand for freight transport as well as the number of overloaded trucks on highways have been increasing in Europe and in other regions of the world (Huiying, et al., 2022; Znidaric, 2015; Mat, et al., 2021; Ullah, et al., 2014). This increase in the number of overloaded trucks on highways can be explained by the fact that in contexts of increasing costs of living and soaring fuel prices, transport enterprises favor overloaded transportation because it enables them to reduce operating costs. However, overloaded trucks have a direct impact on road safety, and can considerably damage road pavements and bridges, thereby reducing their lifespan (Lou, et al., 2017; Ojha, 2018). Moreover, in some cases, they are responsible for tragic accidents (Huiying, et al., 2022; Jacob, et al., 2020; Dontu, et al., 2020). In order to remedy this problem, truck overload control direct enforcement have to be more efficient. The identification and classification of vehicles play a crucial role in ensuring the effective direct enforcement of overloaded vehicle regulations. Current technologies such as weigh-in-motion (WIM) systems can classify most vehicle categories. WIM systems are the most commonly used techniques for the measurement of traffic loading because they can provide a good estimate of the weight of a vehicle (axle weight) without requiring the vehicle to come to a stop. Nevertheless, WIM systems struggle to accurately classify vehicles with lifted axles (see Figure 3) because they rely on force sensors such as piezoelectric sensors, bending plates and single load cells (Dontu, et al., 2020) which can only detect wheels that exert a force on the pavement. Moreover, there are very few commercial and technical methods available that can be used to deduce the position of lifted axles (Marszalek, et al., 2018; Marszalek & Krzysztof, 2020; Quoy & Jacob., 2021; Dimitri, et al., s.d.; Jacquemin & Stanczyk, 1994).

Over the last decades, several image processing methods have been employed for classifying vehicles, ranging from classical image processing methods like the circular Hough transform (Frenzel, 2002; Gothankar, et al., 2019) to more advanced methods employing convolutional neural networks (CNNs) such as Region Based Convolutional Neural Network (R-CNN), Faster R-CNN, You Only Look Once (YOLO), and Single shot multibox Detector (SSD) (Marcomini & Cunha, 2022; Marcomini, et al., 2023; Joseph, et al., 2016). SSD and YOLO are single-stage detectors, i.e., they carry out feature extraction and object detection in a single process, while algorithms like R-CNN and Faster R-CNN carry out feature extraction and object detection in two stages (Marcomini & Cunha, 2022). The two-stage detectors are more precise than single-stage detectors, however, they are not well suited for real-time applications given that they have a higher computational cost. Furthermore, compared to other CNNs in the literature, YOLO offers a good compromise between performance and computational time (Marcomini & Cunha, 2022). To the best of our knowledge, the camera-based methods in the literature focus on the detection of vehicle types (Bipul, et al., 2022; Magdalena, et al., 2020), the number of truck axles (Marcomini & Cunha, 2022; Marcomini, et al., 2023), the estimation of the speed of vehicles (Victoria, et al., 2022), however, none of them deals with the detection of lifted axles.

In this paper, a method based on an advanced version of YOLO, namely YOLOv8s (Jocher, et al., 2023), where the "s" denotes the small variant of YOLOv8, is proposed for the detection of lifted truck axles in images of trucks captured by cameras placed perpendicular to the direction of traffic. The proposed method aims to provide an innovative method for the detection of lifted axles within the scope of SETO (Smart Enforcement of Transport Operations), a European project launched in 2023. SETO aims to create a unified platform that will provide regulatory authorities with real-time access to all necessary information for the smart enforcement of transport and safety regulations.

The rest of this paper is organized as follows. In Section 2, the proposed camera-based method for the detection of lifted truck axles as well as the database used for training and validation are detailed. In Section 3, the performance of the proposed method is assessed. Conclusions and perspectives are presented in Section 4.

## 2. Method

In this work, the method proposed for the detection of lifted truck axles consists of 3 main parts: (I) a truck detection module, (II) an axle detection module, and (III) a lifted axle detection module. Figure 1 shows a schematic representation of the proposed method, with modules for truck detection, axle detection and lifted axle detection shown in blue, orange and purple, respectively. The proposed method first detects a truck and its axles, then uses instance segmentation to identify the pixel masks representing the shapes of any lifted axles. In this paper, all the above-mentioned modules are based on YOLOv8s (Jocher, et al., 2023), with the first two employing the object detection capabilities of YOLOv8s, while the third module employs the instance segmentation capabilities of YOLOv8s. Object detection is a computer vision task that involves predicting bounding boxes around objects and classifying them, while instance segmentation goes further to add a pixel mask that gives the shape of the object for every detected object.

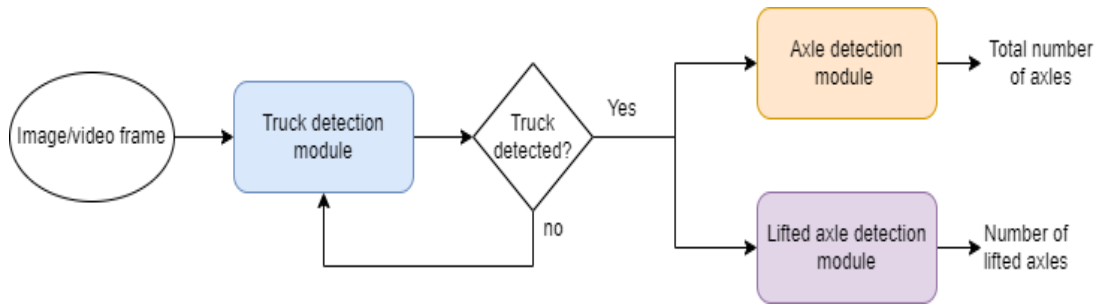

**Figure 1 – Schematic representation of the method proposed for lifted truck axle detection**

### 2.1 Architecture of YOLOv8

YOLOv8 is a cutting-edge version of YOLO (Joseph, et al., 2016) that can carry out object detection, object classification, instance segmentation, tracking and pose estimation (Jocher, et al., 2023) unlike all the previous YOLO versions which were primarily designed for object detection. Five variants of YOLOv8 exist, namely YOLOv8n, YOLOv8s, YOLOv8m, YOLOv8l and YOLOv8x. The suffixes "n", "s", "m", "l", and "x" represent the nano, small, medium, large and extra-large variants, respectively, arranged in order of increasing computational complexity and accuracy (Casas, et al., 2024). In this paper, the proposed method is based on YOLOv8s because research has shown that it delivers higher performance at a lower computational cost compared to YOLOv8n, making it well-suited for real-time implementations (Jocher, et al., 2023).

Similar to previous YOLO versions, the architecture of YOLOv8 consists of 3 parts, namely the backbone, the neck and the head as shown in Figure 2 (Fang, et al., 2021; Casas, et al., 2024; Wenjie, et al., 2023). The backbone is used to extract feature maps from the input image. It extracts different levels of features (or patterns) in an image, ranging from low level features like lines, textures and edges, to high level features like the shape of lifted axles for instance. The backbone of YOLOv8 consists of standard convolution layers (conv), a fast implementation of Cross Stage Partial (CSP) bottleneck with 2 convolutions layers known as C2f, and a Spatial Pyramid Pooling Fusion (SPPF) layer. Conv layers are essential for initial feature extraction. C2f enhances the ability of the model to detect fine details, and SPPF enhances feature extraction by capturing multi-scale information while ensuring fast computation unlike previous YOLO versions (like YOLOv5 (Casas, et al., 2024; Jocher, 2020)) which use less advanced techniques.

The neck is used to enhance the extracted features by combining information from multiple scales using concatenation and up sampling operations (see Figure 2) (Casas, et al., 2024; Wenjie, et al., 2023). The head is responsible of processing the outputs from the neck. It produces final predictions of bounding boxes, classes, and segmentation masks when instance segmentation is performed (Casas, et al., 2024; Wenjie, et al., 2023).

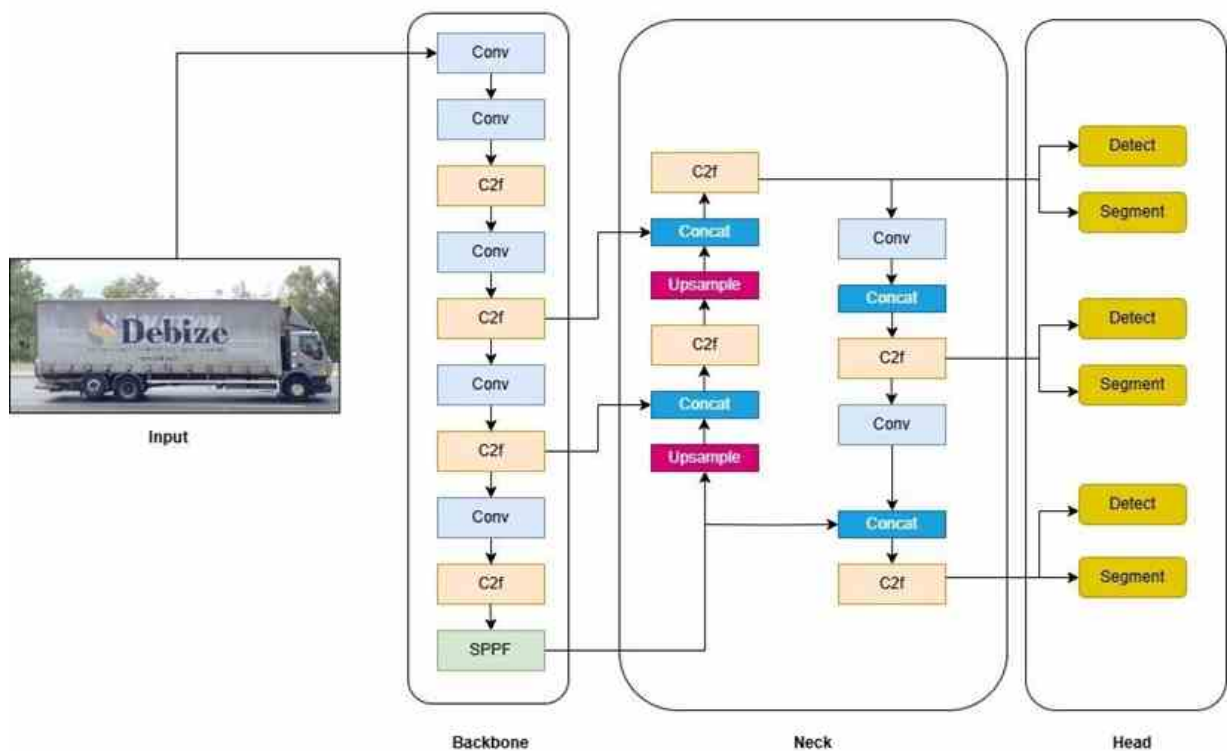

**Figure 2** - **Architecture of YOLOv8 with the detection and segmentation branches in the head**

## 2.2 Dataset description

Two datasets are used in this work. The first dataset, Dataset 1 is used for training and validating the YOLOv8s object detection model employed for truck and axle detection in the proposed method (see the blue and orange modules in Figure 1). It consists of 810 images of trucks featuring both lifted and non-lifted axles, cropped from videos recorded using different cameras placed perpendicular to the direction of traffic. Out of the 810 images found in this dataset, 727 of them were collected in Brazil, in the state of Sao Paulo using 3 different cameras on five different locations (Leandro & Andre, 2021). The dataset in (Leandro & Andre, 2021) is very appealing to the application at hand because it already contains the labels of the trucks and axles present in the images. Nevertheless, it does not contain any labels for lifted axles. The remaining 83 images that constitute Dataset 1 were included in order to increase the number of images featuring trucks with lifted axles. These images were cropped from a video of about 1 hour in length recorded on a sunny day on the A7 highway near Lyon in France, using the main rear camera of a Samsung Galaxy A51 placed at 1.3 m above the ground.

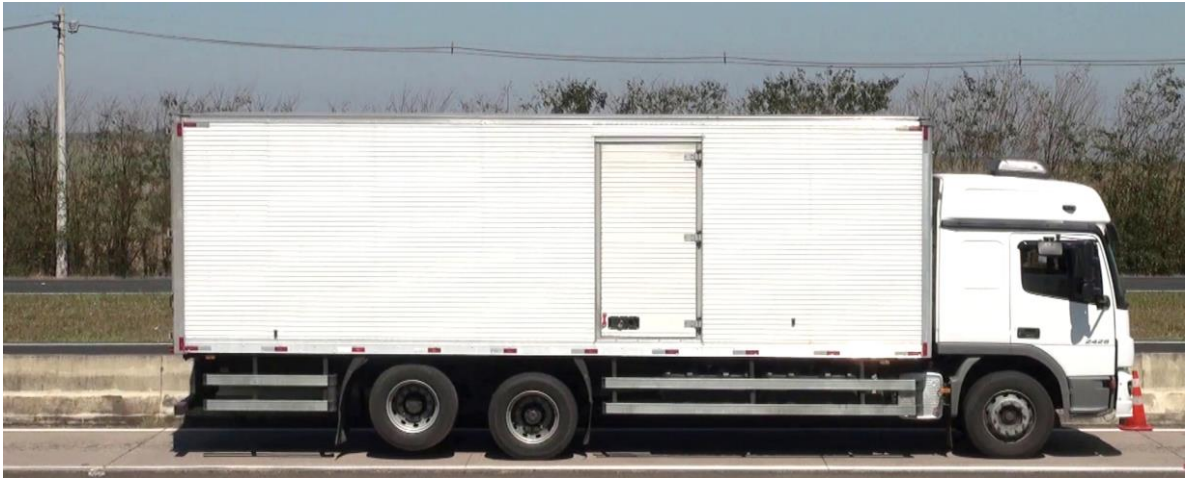

(a) 3-axle truck with 1 lifted axle

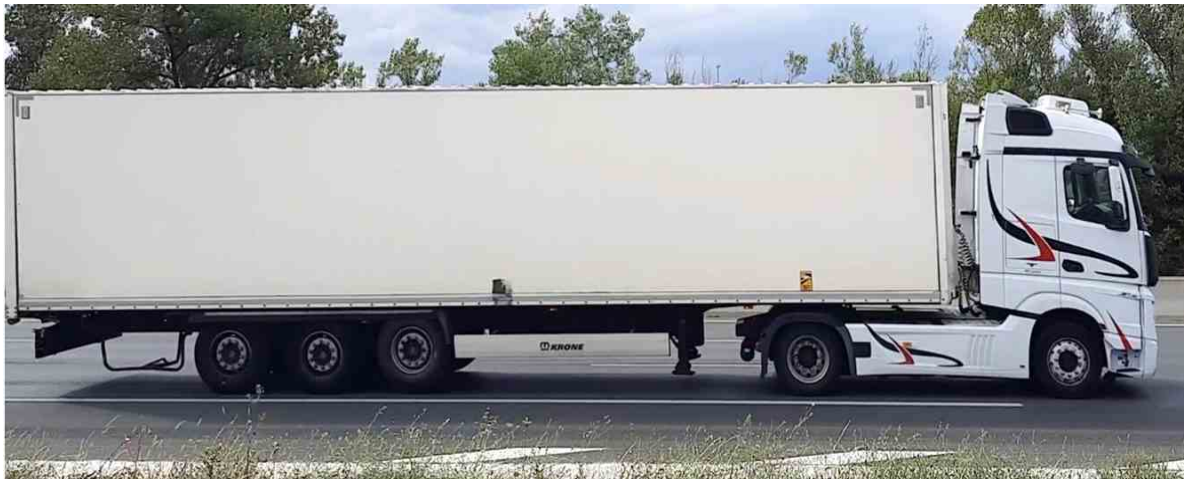

(b) 5-axle truck with 1 lifted axle

**Figure 3 - Examples of trucks with lifted axles found in Dataset 2**

The second dataset, Dataset 2, is a subset of Dataset 1, and it comprises only of images of trucks with lifted axles. It is used to train and validate the YOLOv8s instance segmentation model, i.e., YOLOv8s-seg, employed for predicting the pixel masks of lifted axles (see the purple module in Figure 1). Dataset 2 was formed by selecting images in Dataset 1 in which the lifted axles are clearly visible. It consists of 77 images, of which 52 are from the data collected in Brazil and the remainder are from the data collected in France.

The labels of all the 83 images collected on the A7 highway, as well as the segmentation labels of the lifted axles were produced using a free labelling tool named Makesense AI. Figure 3 shows examples of trucks with lifted axles found in Dataset 2. It can be observed in Figure 3 a and b, that the third axle (counting from the steer axle) is lifted.

Table 1 summarizes the characteristics of the trucks with lifted axles found in Dataset 2. It can be seen that most of the trucks with lifted axles in the images collected in Sao Paulo, and those collected close to Lyon had 3 and 5 axles, respectively.

**Table 1 - Summary of the classes of trucks with lifted axles found in Dataset 2**

| Source | Truck class (number of axles) | | | | | | |
|---|---|---|---|---|---|---|---|
| | 3-axles | 4-axles | 5-axles | 6-axles | 7-axles | 8-axles | 9-axles |
| Brazil (52 trucks) | **20** | 8 | 10 | 11 | 2 | - | 1 |
| France (25 trucks) | 4 | - | **20** | 1 | - | - | - |

## 2.3 Data training

The YOLOv8s object detection and instance segmentation models were trained and validated on Dataset 1 and Dataset 2, respectively, using the official repository of YOLOv8 provided by Ultralytics (Jocher, et al., 2023). Data training was carried out in PyCharm on a computer equipped with an 11 GB NVIDIA GeForce RTX 2080 Ti graphics card. Each dataset was partitioned into training and validation sets in a ratio of 80% to 20%. Each image was resized to 640x640 pixels before processing. The training and validation sets obtained from Dataset 1 consisted of 646 images and 164 images, respectively. For Dataset 2, the training and validation sets consisted of 60 images and 17 images, respectively.

**Table 2 - Hyperparameters used for tunning the YOLOv8s models**

| Hyperparameters | Value in detection model | Value in segmentation model |
|---|---|---|
| Number of epochs | 400 | 400 |
| Batch size | 3 | 32 |
| Optimizer | AdamW | AdamW |
| Learning rate | 0.01 | 0.01 |
| Scaling | 0.5 | 0.5 |
| Flip left-right | 0.5 | 0.5 |
| Shear | 0.5 | - |

During the training process, the weights of the YOLOv8s models were initialized with pretrained weights obtained over the COCO (Common Objects in Context) dataset (Jocher, et al., 2023). This approach was used in order to speed up the training process. Additionally, the AdamW optimizer with a learning rate of 0.01 was used to adjust the weights of the models. Moreover, to improve the models' generalization ability, 3 data augmentation techniques were used: scaling, flipping and shear transformation. During training on Dataset 1, the number of epochs and batch size were set to 400 and 3, respectively, after tuning. With Dataset 2, the number of epochs was the same as with Dataset 1, and the batch size was set to 32. Table 2 presents a summary of the hyperparameters used for tuning the YOLOv8s detection and instance segmentation models.

## 2.4 Evaluation criteria

In this work, the truck, axle, and lifted axle detection modules are assessed using metrics commonly used in the literature (Marcomini & Cunha, 2022; Casas, et al., 2024), namely precision (P), recall (R), F1-score, and mean-average precision (mAP). It is worth mentioning that all these metrics yield values in the range [0, 1] or equivalently, 0% to 100%. Additionally, both the training time and the inference time are measured, with the latter indicating the time required by a model to detect or segment objects of interest in an image. Furthermore, confusion matrices are computed in order to evaluate the classification capabilities of YOLOv8s and YOLOv8s-seg.

Precision refers to how many of the predicted bounding boxes correctly match (or overlap with) actual objects. The overleap between the predicted and the ground truth bounding boxes is measured by the Intersection over Union (IoU):

$$\text{IoU} = \frac{\text{Area of overlap}}{\text{Area of Union}}.$$ (1)

The IoU threshold determines if a predicted box is true positive or not. For instance, if the IoU threshold is greater than or equal to 0.5, and a predicted box's IoU with a ground truth is greater than or equal to 0.5, then the prediction is considered a true positive. Precision is computed as follows:

$$\text{Precision} = \frac{\text{TP}}{\text{TP} + \text{FP}}$$ (2)

where TP and FP refer to true positives and false positives, respectively. A high precision implies that the detection and instance segmentation models are confident and accurate about the detections made (Casas, et al., 2024; Marcomini & Cunha, 2022). Recall (also known as sensitivity) represents the fraction of correctly detected objects among all actual instances (or occurrences) of that object in the dataset. A high recall score indicates that a model is good at finding the objects under study. Recall is calculated as:

$$\text{Recall} = \frac{\text{TP}}{\text{TP} + \text{FN}}$$ (3)

where FN refers to false negatives.

The F1-score is the harmonic mean of precision and recall (Marcomini & Cunha, 2022; Casas, et al., 2024). It provides a single metric that balances the trade-off between precision and recall. A high F1-score indicates that a model has a good overall performance. The F1-score is computed as:

$$\text{F1-score} = \frac{2(\text{Precision x Recall})}{\text{Precision} + \text{Recall}}.$$ (4)

The mAP is the mean of the Average Precision (AP) values computed for all object classes in the datasets under study. For Dataset 1, the mAP is computed over two classes, namely, truck and axle, while for Dataset 2, it is computed over a single class, namely lifted axle. The mAP is computed as:

$$mAP = \frac{1}{n} \sum_{k=1}^{n} AP_k \tag{5}$$

where n is the number of classes. In this work, the mAP is calculated at two different IoU thresholds, $IoU \geq 0.5$ and $0.5 \leq IoU \leq 0.95$. The former and the latter are commonly written as mAP50 and mAP50-95, respectively.

## 3. Results and discussions

In this section, the performance of the proposed method is assessed using Dataset 1 and 2.

### 3.1 Truck and axle detection: Performance assessment on Dataset 1

Table 3 presents the best performance of YOLOv8s obtained following the completion of the training and validation processes on Dataset 1 (refer to Section 2). The performance metrics are averaged over two classification classes: truck and axle. The precision and recall metrics of YOLOv8s model on both the training and validation sets are notably high, around 99%. This performance suggests that the majority of the detected trucks and axles are indeed true positives. Moreover, YOLOv8s has a good balance between recall and precision as evidenced by its high F1-score of about 98.8 % on the training and validation sets. In terms of mAP50 and mAP50-95, YOLOv8s achieves scores of about 99.1% and 90.1% respectively, on both the training and validation sets. These high mAP values indicate that YOLOv8s achieves high precision at various IoU thresholds. The training time (on 646 images) and the inference time were 1.5 hours and 2.0 ms, respectively.

**Table 3 – Performance of the YOLOv8s model used for truck and axle detection**

| YOLO model | Precision | | Recall | | F1-score | | mAP50 | | mAP50-95 | |
|---|---|---|---|---|---|---|---|---|---|---|
| | Train. | Val. | Train. | Val. | Train. | Val. | Train. | Val. | Train | Val. |
| 8s | 0.9904 | 0.9900 | 0.9854 | 0.9850 | 0.9879 | 0.9875 | 0.9908 | 0.9910 | 0.9007 | 0.9010 |

The confusion matrix in Figure 4 gives more details concerning the performance of YOLOv8s in classifying instances of trucks and axles in the validation set. The background class refers to any area in the image that does not contain any objects of interest. It can be seen that out of the 167 instances of trucks in the validation set, 164 of them were correctly predicted as trucks (true negatives), while the remaining 3 were incorrectly predicted as background (false negatives). Thus, the recall (or percentage of correctly detected trucks) is 98.2%.

Furthermore, it can be observed that there is a total of 623 instances of axles in the validation set. YOLOv8s correctly predicted 618 of them as axles, and incorrectly predicted the remaining 5 as background, yielding a recall 99.2%. The last column of the confusion matrix shows the number of instances during which the background was incorrectly classified as truck and axle.

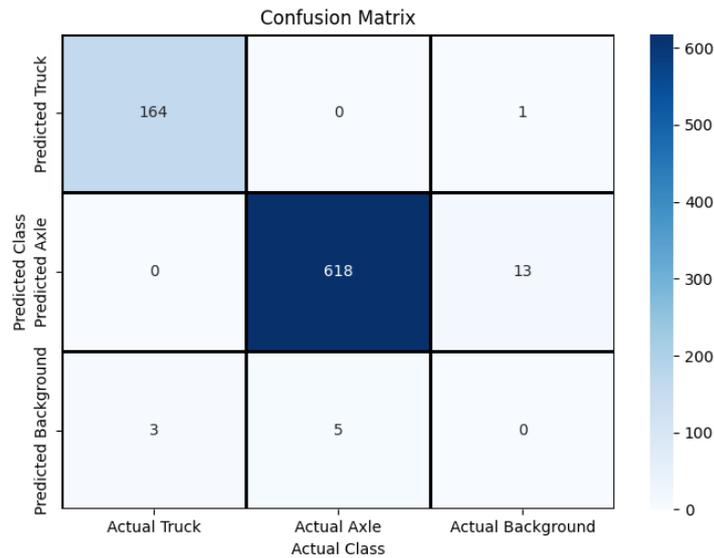

**Figure 4 - Confusion matrix obtained on Dataset 1**

Figure 5 shows the detection results obtained after applying the trained YOLOv8s model on the 5-axle truck in Figure 3 b. It can be seen that the truck and its axles are detected with confidence scores greater than or equal to 90%.

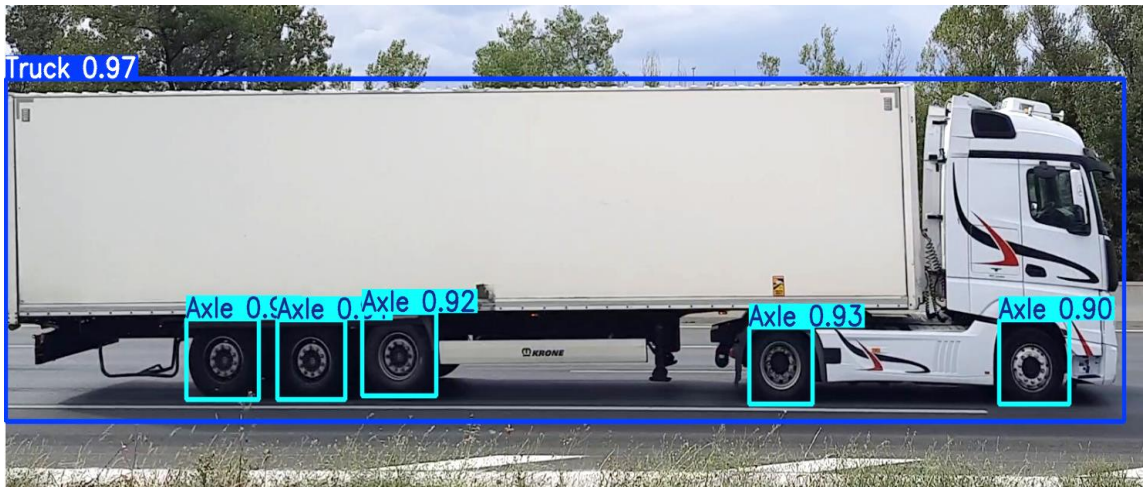

**Figure 5 – Example of truck and axle detections**

### 3.2 Lifted axle detection: performance assessment on Dataset 2

Table 4 displays the best performance of YOLOv8s-seg obtained after completing the training and validation processes on Dataset 2. The model's performance was evaluated for a single classification class, namely lifted axle. It can be seen that the precision and recall scores of YOLOv8s-seg are high, achieving scores of 87.0% and 87.5 %, respectively, on both the training and validation sets. In terms of F1-score, YOLOv8s-seg achieves a score of 82.3% on both the training and validation sets. This high F1-score indicates that YOLOv8s-seg is well suited for accurately detecting lifted axles while minimizing both false positives and false negatives. Furthermore, YOLOv8s-seg achieves mAP50 and mAP50-95 scores of about 91.8% and 82.3%, respectively, on both the training and validation sets. The training time on 60 images was found to be 7 minutes, and the inference time was 1.4 ms.

**Table 4 - Performance of YOLOv8s-seg on Dataset 2**

| YOLO model | Precision | | Recall | | F1-score | | mAP50 | | mAP50-95 | |
|---|---|---|---|---|---|---|---|---|---|---|
| | Train. | Val. | Train. | Val. | Train. | Val. | Train. | Val. | Train | Val. |
| 8s-seg | 0.8702 | 0.8700 | 0.8750 | 0.8750 | 0.8726 | 0.8725 | 0.9189 | 0.9190 | 0.8232 | 0.8250 |

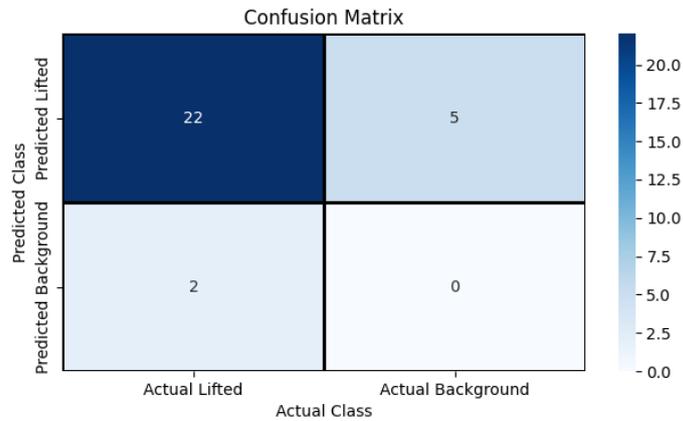

**Figure 6** - **Confusion matrix obtained on Dataset 2**

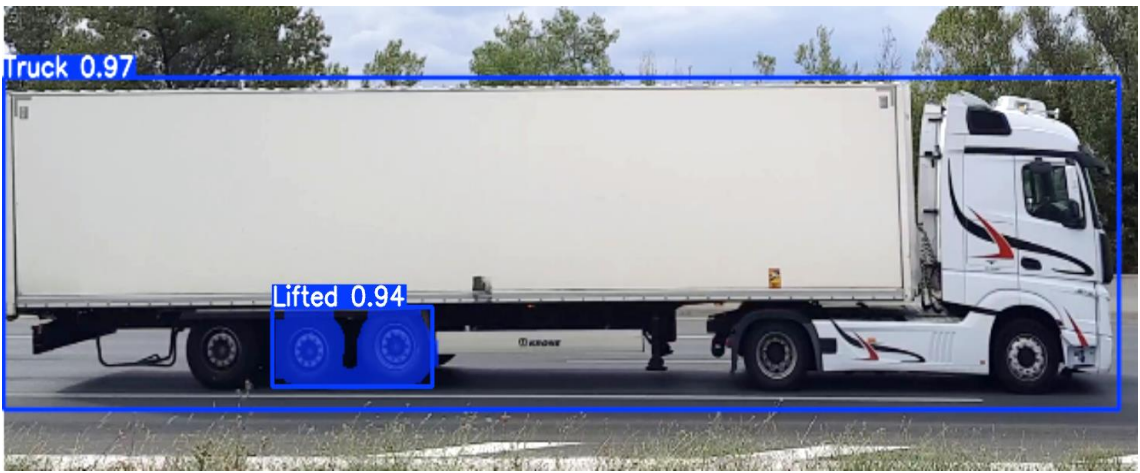

**Figure 7 – Example of lifted axle detection by the method**

Figure 6 displays the confusion matrix obtained after employing YOLOv8s-seg on the validation set. It can be observed that out of the 24 instances of lifted axles in the validation set, 22 of them were correctly predicted as lifted axles, while the remaining 2 were incorrectly predicted as background. This results in a recall of 91.7%. Notably, the difference between this recall and the value reported in Table 4 is about 4.2%, which may be attributed to differences in the IoU threshold used during evaluation. Figure 7 shows the results obtained after applying YOLOv8s-seg on the 5-axle truck in Figure 3. The predicted bounding box and mask clearly show the lifted axle, with a high confidence score of 94%.

## 4. Conclusions

In this paper, a method for detecting lifted axles in images of trucks recorded with cameras positioned perpendicular to the direction of traffic was proposed. The proposed camera-based method employed the object detection and instance segmentation capabilities of YOLOv8s, a cutting-edge CNN to predict the shape of lifted axles in cropped images. The performance of the method was assessed using metrics commonly used in the literature and it was found that it had a precision of 87%, a recall (or percentage of correctly detected lifted axles) of 91.7% and an inference time of 1.4 ms, which makes it well-suited for real time implantations. Future work will focus on improving the performance of the proposed method by increasing the size of the datasets using images recorded by different cameras, and by exploring additional image augmentation techniques. Moreover, the effectiveness of other YOLOv8 variants and SSD will be examined to identify potential improvements.

## 5. Acknowledgements

This research was funded by the Smart Enforcement of Transport Operations (SETO) project.